\pdfoutput=1

\documentclass[11pt]{article}

\usepackage[final]{acl}
\usepackage{tablefootnote}
\usepackage{times}
\usepackage{latexsym}
\usepackage{threeparttable}

\usepackage[T1]{fontenc}
\usepackage{soul} 

\usepackage{color, xcolor}

\usepackage[utf8]{inputenc}
\usepackage{enumitem}
\setenumerate[1]{itemsep=1.5pt,partopsep=0pt,parsep=\parskip,topsep=5pt}
\setitemize[1]{itemsep=1.5pt,partopsep=0pt,parsep=\parskip,topsep=5pt}
\usepackage{microtype}

\usepackage{inconsolata}

\usepackage{graphicx}

\title{MedExQA: Medical Question Answering Benchmark \\with Multiple Explanations}

\author{Yunsoo Kim, Jinge Wu, Yusuf Abdulle, Honghan Wu \\
  Institute of Health Informatics, University College London \\
  \texttt{yunsoo.kim.23@ucl.ac.uk} \\
  }
\begin{document}
\maketitle
\begin{abstract}
This paper introduces MedExQA, a novel benchmark in medical question-answering, to evaluate large language models’ (LLMs) understanding of medical knowledge through explanations. By constructing datasets across five distinct medical specialties that are underrepresented in current datasets and further incorporating multiple explanations for each question-answer pair, we address a major gap in current medical QA benchmarks which is the absence of comprehensive assessments of LLMs’ ability to generate nuanced medical explanations. Our work highlights the importance of explainability in medical LLMs, proposes an effective methodology for evaluating models beyond classification accuracy, and sheds light on one specific domain, speech language pathology, where current LLMs including GPT4 lack good understanding. Our results show generation evaluation with multiple explanations aligns better with human assessment, highlighting an opportunity for a more robust automated comprehension assessment for LLMs. To diversify open-source medical LLMs (currently mostly based on Llama2), this work also proposes a new medical model, MedPhi-2, based on Phi-2 (2.7B). The model outperformed medical LLMs based on Llama2-70B in generating explanations, showing its effectiveness in the resource-constrained medical domain. The benchmark datasets and the model can be found at \url{https://github.com/knowlab/MedExQA}.
\end{abstract}

\section{Introduction}
\label{sec:intro}
Recent advancements in large language models (LLMs) have not only enhanced their understanding of medical domain text but also improved their ability to generate coherent text with correct medical knowledge \cite{tu2023towards,singhal2023towards}. Chatbots, powered by these LLMs, have emerged as indispensable tools, offering unprecedented opportunities to enhance patient care, streamline clinical decision-making processes, and medical knowledge retrieval \cite{achiam2023gpt, chatgpt, grovesbiomedknowledge}. Moreover, open-source medical LLMs further enhance the usability of such technologies in hospitals by resolving the privacy concerns associated with patient data \cite{toma2023clinical,kweon2023publicly,chen2023meditron}.

This research in medical LLMs has been facilitated by the introduction of question-answering (QA) datasets that serve as benchmarks for evaluating the model's understanding of medical domain knowledge \cite{hendrycks2020measuring, jin2021disease, pal2022medmcqa, singhal2023towards}. The benchmark QA datasets typically consist of multiple-choice questions (MCQ), enabling researchers to readily assess the capabilities of LLMs in comprehending and responding to diverse medical inquiries. Thus, the diversity within these datasets is a key component in creating a rigorous assessment benchmark for complex medical concepts. Nonetheless, certain areas within the medical domain, such as speech language pathology, still remain uncovered by the current benchmark datasets.

As current medical QA benchmarks are often structured as MCQ, classification accuracy is used as an evaluation metric. However, classification accuracy alone may not adequately assess whether LLMs possess the nuanced medical expertise required for reasoned responses. The explanation and rationale behind the selection of a particular choice by an LLM would provide a deeper understanding of the model's capabilities and limitations in generating responses to intricate medical questions. This comprehensive evaluation, delving into the explanation and rationale, is especially important in clinical settings where misleading information such as hallucinations produced by LLMs can have serious consequences. 


In order to assess the quality of the model explainability, the dataset should include a golden explanation for the reasoning behind the answer. Additionally, since there are often multiple ways to express the same rationale in text, an ideal dataset would provide a multiple set of explanations for a single QA pair. However, current benchmark datasets are not focused on providing explanations as they often lack explanations entirely or only a subset of the dataset comes with an explanation \cite{hendrycks2020measuring, jin2021disease, pal2022medmcqa}. This limitation highlights the need for improved datasets that are explicitly designed to include comprehensive explanations.

To address this issue, this paper presents a novel QA benchmark, MedExQA, with two sets of explanations, aiming to provide a more comprehensive evaluation of LLMs in the medical domain. To diversify the knowledge coverage in the current datasets, our proposed benchmark consists of five underrepresented specialties in current datasets: biomedical engineering, clinical laboratory science, clinical psychology, occupational therapy, and speech language pathology. In this work, the datasets were used to benchmark the performance of an extensive list of LLMs, including those trained with medical domain text. With this comprehensive benchmark evaluation, we explored the effects of medical domain-specific training. Additionally, to diversify the pool of open-source medical LLMs which are currently almost all based on the Llama2 model, we introduce our own trained model, MedPhi-2, a Phi-2 model trained with medical domain text. Our MedPhi-2 model outperformed medical LLMs based on the Llama2-70B model in generating explanations for the rationale behind the answer.

The contributions of this paper are as follows:
\begin{enumerate}
    \item \textbf{MedExQA novel datasets with explanations.} We constructed a benchmark with 5 distinct specialties within the medical domain. The datasets include two explanations for each question and answer pairs. 
    \item \textbf{Comprehensive Benchmark.} We evaluated an extensive list of models: 18 baseline open-source models with various sizes (from 2.7B to 70B), 3 OpenAI GPT models, as well as our model (detailed below). In terms of evaluation approach, classification accuracy, generated explanation performance, and human evaluations are considered. To highlight, this is the first benchmark using multiple explanations, and the results demonstrate that our benchmark can better evaluate language models' understanding of medical domain knowledge.
    \item \textbf{MedPhi-2 model.} We trained a small language model (SLM) based on the Phi-2 model, with medical pretraining corpus and instruction-tuning datasets. The model outperformed medical LLMs based on Llama2 70B in generating explanations. 
    \item \textbf{Open source.} We release the datasets, model weights, and codes to facilitate the research in medical large language modeling.
\end{enumerate}

\section{Related Works}
\label{sec:related_work}

\subsection{MMLU}
\label{sec:rel_mmlu}
MMLU \cite{hendrycks2020measuring} is a benchmark designed to measure the model's ability in knowledge-intensive QA with four-way MCQs. Within the extensive list of subjects, there are nine healthcare-related subjects such as professional medicine and medical genetics. Collectively, these nine subjects comprise a total of 1,871 questions in the test set. While MMLU provides a comprehensive set of questions, it lacks explanations for the answers, thereby limiting the dataset's evaluation to mere multiple-choice classification accuracy.

\subsection{MedQA}
\label{sec:rel_medqa}
MedQA \cite{jin2021disease} is an open-ended MCQ dataset made from professional medical doctor license exams. The dataset contains questions drawn from both real exams and mock tests for the United States Medical License Exams \cite{usmle}. 1,273 questions, each question accompanied by four or five answer choices, are provided as the test dataset. Similar to MMLU, MedQA does not include explanations for assessing the ability to generate rationale behind the answer.

\subsection{MedMCQA}
\label{sec:rel_medmcqa}
MedMCQA \cite{pal2022medmcqa} is a benchmark with questions sourced from postgraduate-level Indian medical school entrance exams (AIIMS and NEET PG). The dataset covers a breadth of medical specialties, 2,400 healthcare topics and 21 subjects and provides 4,183 MCQ with four answer choices for evaluation. Although MedMCQA is known to have explanations, nearly half of the evaluation dataset lacks explanations and instances of duplicate explanations are also observed. In fact, accuracy is only reported as the evaluation metric and explanation is not used in their paper entirely. Therefore, MedMCQA is not primarily designed for the assessment of generating explanations.

\section{MedExQA Datasets}
We introduce MedExQA, a novel QA benchmark designed to tackle the limitations of existing benchmarks by incorporating two sets of explanations. This approach aims to offer a more thorough evaluation of performance in five underrepresented specialties in the medical domain: Biomedical Engineering, Clinical Laboratory Science, Clinical Psychology, Occupational Therapy, and Speech Language Pathology.

\label{sec:medexqa}
\subsection{Datasets Preparation}
The raw data was manually collected from diverse freely accessible online sources, including mock tests and online exams tailored to each medical professional specialty. Some questions of the mock tests and online exams have explanations for the answers, which we used the creation of the MedExQA datasets. The pass mark for the collected mock tests and online exams was 60 percent.

To ensure data integrity, rigorous preprocessing was conducted, including the removal of duplicate questions and explanations. Additionally, similar questions were identified and eliminated using BERT cosine similarity analysis \cite{devlin2018bert}. Questions containing keywords specific to laws or regulations were filtered out using a manually curated list of words. Following fair use regulations\footnote{https://www.copyright.gov/fair-use/more-info.html}, answer options were systematically shuffled to maintain fairness and uphold the integrity of the dataset.
Furthermore, to enhance the quality and coherence of the datasets, two sets of explanations as well as the questions underwent thorough human validation. This validation process aimed to ensure that the explanations exhibited distinct writing styles and provided comprehensible reasoning for the correct answer selection.

\begin{table}
\centering
\begin{tabular}{lll}
\hline
\textbf{Specialty} & \textbf{NUM} &\textbf{ExSIM}\\
\hline
Biomedical Engineering & 148 & 75.8\\
Clinical Laboratory Science & 377 & 73.7\\
Clinical Psychology & 111 & 79.7\\ 
Occupational Therapy & 194 & 79.5\\ 
Speech Language Pathology & 135 & 80.5\\\hline
\textbf{Total} & 965  & 78.7\\\hline
\end{tabular}
\caption{Statistics of datasets within \textbf{MedExQA}. \textbf{NUM} represents the number of questions. \textbf{ExSIM} represents the average cosine similarity of the explanation pairs. }
\label{tab:num_ques}
\end{table}

The resulting datasets have a total of 965 questions. Table \ref{tab:num_ques} provides a detailed breakdown of the number of questions for each specialty. These datasets were split into a few-shot development set and a test set. Specifically, the few-shot development set has 5 questions per specialty, while the test set consists of 940 questions in total. It is noteworthy that each subject contains a minimum of 100 test examples, a length surpassing that of most exams tailored for human assessment.

Also, to validate that each pair of explanations is different sufficiently at the individual question level, Table \ref{tab:num_ques} also provides the average cosine similarity of the pairs. The overall similarity is 78.7\% which indicates the lexical difference of the two corresponding versions of explanations for each question. An example of the dataset as well as the difference in the pair of example can be found in the Appendix Figure \ref{fig:exp}.


\begin{figure}[htbp]
    \centering
    \includegraphics[trim={0cm 0cm 0cm 0cm},clip,width=0.48\textwidth]{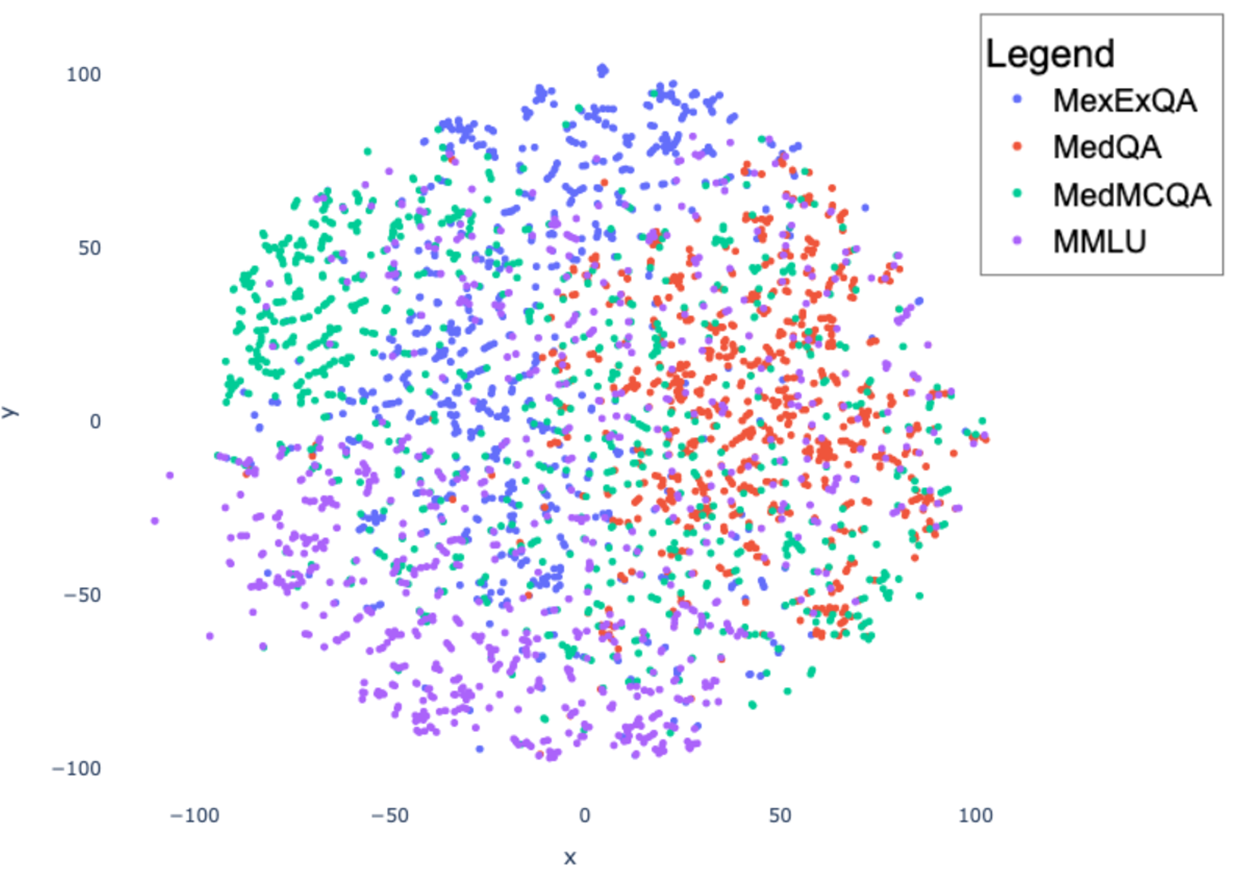}
    \caption{2D t-SNE plot for MedExQA, MedQA, MedMCQA, and MMLU (Medicine Related 9 subjects) datasets.}
    \label{fig:tsne}
\end{figure}

\subsection{Comparison of benchmark datasets}
We compared MedExQA with existing benchmark datasets by visualizing their questions in the same vector space. Using t-distributed Stochastic Neighbour Embedding (tSNE), each question is represented as a point in the vector space. We used the `all-mpnet-base-v2' sentence transformer model in \texttt{sklearn} package \texttt{tSNE} to retrieve vectors from questions. 965 questions were randomly sampled from each dataset. There is a cluster towards the top region mainly composed of questions from MedExQA, which clearly demonstrates its novelty compared to existing medical QA datasets.


\section{Methods}
For all the experiments in this paper, both training and evaluation, we used 8 A6000 GPUs. 

\subsection{Baseline Models}
We explored 18 baseline models with different sizes from 2.7B to 70B. Table \ref{tab:models} provides a comprehensive overview of the baseline models used in this paper, while more detailed descriptions of each model are available in the appendix. In cases where multiple sizes of a model are used, we distinguish each version by appending the model size to the model name. For example, the Llama2 models with sizes 7, 13, and 70B are denoted as \textbf{Llama2-7B}, \textbf{Llama2-13B}, and \textbf{Llama2-70B}, respectively. On the other hand, when a model has only one size, we refer to it solely by its name. For instance, \textbf{ClinicalCamel} denotes the ClinicalCamel 70B model.

\begin{table}[htbp]
\centering
\begin{tabular}{p{5.0cm}p{1.85cm}}
\hline
\textbf{Llama2 Variant Models} & \textbf{Model Size}\\
\hline
Llama2\cite{touvron2023llama} & 7B,13B,70B\\
ClinicalCamel\cite{toma2023clinical} & 70B \\
Asclepius\cite{kweon2023publicly} & 7B,13B \\
Med42\cite{med42} & 70B \\
AlpaCare\cite{zhang2023alpacare} & 7B,13B \\
Meditron\cite{chen2023meditron} & 7B,70B \\
Medinote\cite{medinote} & 7B,13B \\
\hline
\textbf{Other foundational models} & \textbf{Model Size}\\
\hline
Mistral\cite{jiang2023mistral} & 7B \\
Yi\cite{yi6b} & 6B \\
Phi-2\cite{phi2} & 2.7B \\
SOLAR\cite{kim2023solar} & 10.7B \\
InternLM2\cite{internlm} & 7B 
\end{tabular}
\caption{Baseline Models. The models are sorted in the order of release dates.}
\label{tab:models}
\end{table}

\subsection{Training MedPhi-2}

As far as we know, all the publicly available open-source medical LLMs are based on Llama models, we further extended our work to test the effect of medical domain training on a different foundational model. Phi-2 model was further trained using the medical datasets that are publicly available. We pretrained Phi-2 with a 110M medical-related corpus. We further finetuned the continued pretrained model with 239K instructions. We refer to the resulting model as \textbf{MedPhi-2} throughout our paper. Table \ref{tab:training_data} summarizes the detailed composition of our training dataset. We used LLaMaFactory\footnote{https://github.com/hiyouga/LLaMA-Factory} and used Deep3 for efficient training. For both pretraining and finetuning, We trained the model with a batch size of 16 and a learning rate of 1e-5 with 3 epochs, which took 36 hours in total.

\begin{table}[htbp]
\centering
\begin{tabular}{ll}
\hline
\textbf{Pretrain} & \textbf{Tokens}\\
\hline
Meditron Medical Guidelines\tablefootnote{https://huggingface.co/datasets/epfl-llm/guidelines} & 48.3M \\
SNOMED CT descriptions\tablefootnote{https://huggingface.co/datasets/FremyCompany/AGCT-Dataset} & 28.3M \\ 
Biomedical Article Abstracts\tablefootnote{https://huggingface.co/datasets/paniniDot/sci\_lay} & 13.6M \\ 
Wikipedia Medical Terms\tablefootnote{https://huggingface.co/datasets/gamino/wiki\_medical\_terms} & 13.3M \\ 
PMC Patients Notes\tablefootnote{https://huggingface.co/datasets/zhengyun21/pmc-patients} & 6.7M \\
\hline
\textbf{Finetuning} & \textbf{Instructions}\\
\hline
Asclepius Instruction\tablefootnote{https://huggingface.co/datasets/starmpcc/asclepius-synthetic-clinical-notes}& 158,114 \\
AlpaCare Instruction\tablefootnote{https://huggingface.co/datasets/casey-martin/medinstruct} & 52,002 \\
NHS QA and Medical Task\tablefootnote{https://github.com/CogStack/OpenGPT} & 29,354 
\end{tabular}
\caption{MedPhi-2 training data. The number of tokens for pretraining data and the number of instructions for finetuning data are listed.}
\label{tab:training_data}
\end{table}

\subsection{Evaluation}

We evaluated all models with test datasets except for human evaluation, which was performed on the development datasets. For all the evaluations, we used zero-shot, a batch size of 1, temperature of 0. To benchmark the performance of closed source models we further extended to include OpenAI's GPT models. We used \texttt{GPT3.5\_1106},  \texttt{GPT4.0\_1106}, and  \texttt{GPT4.0\_0125} APIs\footnote{https://platform.openai.com/docs/models}. 

\subsubsection{Classification Accuracy - Logits}
Classification accuracy of MCQ for generative models relies on classifying the next token using logits. In other words, the token with the highest logit value is selected as the model's predicted answer. However, this approach cannot assess the model's understanding of the rationale behind the answer. We exclude GPT models for this evaluation, as we are not able to get the logit value for the next token.

\subsubsection{Classification Accuracy - Chat}
We utilize string-matching using regular expressions and \texttt{thefuzz} package to assess the model's proficiency in generating accurate textual responses. This approach involves searching the specific phrase for the answer choice or the choice letter within the generated response, enabling a more realistic evaluation for the model's performance.

\subsection{Explanation Generation}
The quality of generated explanations is further assessed using a combination of general lexical metrics. \textbf{BLEU \cite{papineni2002bleu}.} measures the geometric mean of precision scores of the generated explanations compared to reference explanations based on n-gram matches. \textbf{ROUGE \cite{lin2004rouge}.} assesses the similarity between generated and reference explanations, with ROUGE-L, providing a score that combines precision and recall based on the longest common subsequence. \textbf{METEOR \cite{banerjee2005meteor}.} considers the semantic similarity and lexical variations with WordNet. \textbf{BERTScore \cite{zhang2019bertscore}.} uses contextual embeddings, scibert embedding \cite{beltagy2019scibert} for our work, to capture nuances in the semantics of the explanations. All the metrics are calculated using \texttt{evaluate} package. 

We propose an enhanced methodology for evaluating models' understanding of medical domain knowledge by incorporating classification accuracy based on string matches into calculating these metrics. We assign a score of 0 to responses with incorrect answers based on string-matching classification results.

\subsection{Evaluation - Human Evaluation}
For human evaluation, three human annotators with MSc degrees in health-related subjects participated in assessing the quality of generated explanations. The evaluation process involved assigning a score for each explanation-answer pair based on the following rules: 
\begin{enumerate}
    \item \textbf{Score 0} the answer was incorrect, no explanation was provided, and/or the explanation is fully irrelevant.
    \item \textbf{Score 0.5} the answer was correct, but the explanation or rationale was incorrect. Also, an incomplete explanation that ended with an incomplete sentence.
    \item \textbf{Score 1.0} when both the answer and explanation were correct.
    
\end{enumerate}

Although this human evaluation was performed on a small scale (development dataset: 5 samples for each specialty), this systematic evaluation process ensured a comprehensive assessment of the models' performance in providing accurate and coherent explanations.

\section{Results and Discussion}
\label{sec:bibtex}

\begin{table*}[]
\begin{center}
\begin{tabular}{ccccccc}
\hline
\textbf{Model}    & \textbf{BE} & \textbf{CP} & \textbf{SLP} & \textbf{OT} & \textbf{CLS} & \textbf{MAvg} \\ \hline
Medinote-7B                   & 33.6 (-4.9)                          & 34.9 (-8.5)                            & 23.1 (6.2)                                  & 38.1 (-8.5)                             & 44.6 (-11.6)                                   & 34.9 (-5.5)                   \\
Meditron-7B                     & 37.8 (-7.7)                          & 46.2 (-16.0)                            & 20.8 (2.3)                                  & 42.9 (-10.6)                            & 43.3 (-6.7)                                    & 38.2 (-7.8)                  \\					
Llama2-7B                        & 42.0 (-9.1)                          & 47.2 (-9.4)                            & 22.3 (1.5)                                  & 40.2 (-12.7                             & 47.6 (-17.5)                                    & 39.9 (-9.4)                    \\
Asclepius-7B                 & 44.8 (-11.2)                          & 47.2 (-17.0)                            & 27.7 (-1.5)                                  & 42.9 (-15.3)                             & 45.2 (-13.4)                                   & 41.5 (-11.7)         \\
Medinote-13B                       & 46.2 (-18.9)                          & 52.8 (-30.2)                            & 28.5 (-4.6)                                  & 49.2 (-28.1)                             & 52.4 (-20.2)                                    & 45.8 (-20.4)               \\
AlpaCare-7B                  & 53.2 (6.3)                          & 53.8 (1.9)                            & 26.9 (6.2)                                  & 59.8 (-3.7)                            & 54.6 (-0.5)                                   & 49.6 (2.0)    \\					
Asclepius-13B                  & 57.3 (-21.0)                          & 56.6 (-33.0)                             & 25.4 (-3.8)                                  & 59.8 (-34.4)                             & 56.5 (-22.9)                                    & 51.1 (-23.0)             \\					
Phi-2                         & 61.5(-35.7)                         & 68.9 (-38.7)                            & 26.2 (2.3)                                  & 64.0 (-43.4)                             & 50 (-25.0)                                      & 54.1 (-28.1)                  \\
Llama2-13B                         & 63.6 (-26.6)                          & 65.1 (-42.8)                           & 27.7 (16.2)                                  & 60.9 (-28.8)                             & 59.4 (-17.5)                                    & 55.3 (-19.9)                \\
MedPhi-2       & 65.7 (-5.6)                          & 70.8 (0.0)                            & 23.1 (0.0)                                  & 65.1 (-0.5)                             & 55.1 (5.1)                                    & 56.0 (-0.2)       \\
AlpaCare-13B                 & 67.1 (-4.9)                          & 69.8 (-10.4)                            & 26.9 (-1.5)                                  & 65.1 (-4.8)                             & 61.6 (-4.3)                                    & 58.1 (-5.2)         \\
Mistral                           & 75.5 (-11.2)                         & 73.6 (-10.4)                           & 32.3 (-6.2)                                 & 75.7 (-6.3)                            & 71.2 (0.0)                                   & 65.7 (-6.8)         \\
Meditron-70B                        & 78.3 (-36.4)                          & 84.9 (-43.4)                   & 30.8 (-5.4)                                  & 69.8 (-37.0)                             & 68.6 (-24.2)                                    & 66.5 (-29.3)          \\
Yi                               & 75.5 (-20.3)                          & 83.0 (-28.3)                           & 30.8 (0.8)                                  & 74.1 (-20.6)                             & 73.4 (-17.2)                                    & 67.4 (-17.1)               \\
SOLAR                          & 74.8 (0.0)                          & 81.1 (-2.8)                            & \textbf{33.1} (-7.7)                         & 73.0 (-1.1)                             & 76.1 (-3.2)                                    & 67.6 (-3.0)           \\
InternLM2                         & 77.6 (-25.2)                          & 82.1 (-38.7)                            & 29.2 (-5.4)                                  & 74.6 (-36.0)                             & 75.0 (-33.6)                                       & 67.7 (-27.8)                \\
ClinicalCamel                   & 78.3 (-6.3)                          & 84.0 (-14.1)                            & 28.5 (-5.4)                                  & 79.9 (-6.3)                             & 75.8 (-6.2)                                    & 69.3 (-7.7)               \\
Llama2-70B                         & 78.3 (-10.5)                          & 84.0 (-47.2)                            & 31.5 (-10.8)                                  & 80.4 (-44.4)                    & 72.9 (-29.8)                                    & 69.4 (-28.5)            \\
Med42                            & 83.2 (-14.)                 & 84.9 (-10.4)                  & 31.5 (-4.6)                                  & 79.4 (-13.8)                             & 80.9 (-12.6)                           & 72.0 (-11.1)  \\
GPT3.5\_1106                   & 72.0       & 82.1       & 29.2        & 70.4       & 71.5        & 65.0            \\
GPT4\_1106                      & 86.7       & 86.8       & 31.5        & 88.4       & \textbf{91.7}        & 77.0               \\
GPT4\_0125                   & \textbf{90.2}       & \textbf{91.5}       & 30.8        & \textbf{90.0}       & \textbf{91.7}        & \textbf{78.8}              \\
\hline
\end{tabular}
\caption{\label{tab:result_choice}
MCQ accuracy (\%) using logits vs chat generation. The MCQ accuracy using logits is reported (except for GPT models). The performance gain/loss with chat generation approach is marked in parenthesis. "BE": Biomedical Engineering; "CP": Clinical Psychology; "SLP":	Speech Language Pathology; "OT":	Occupational Therapy;	"CLS": Clinical Laboratory Science; "MAvg": Macro Average.
}
\end{center}
\end{table*}

\begin{figure*}[htbp]
    \centering
    \includegraphics[trim={0cm 4cm 0cm 6.7cm},clip,width=1.\textwidth]{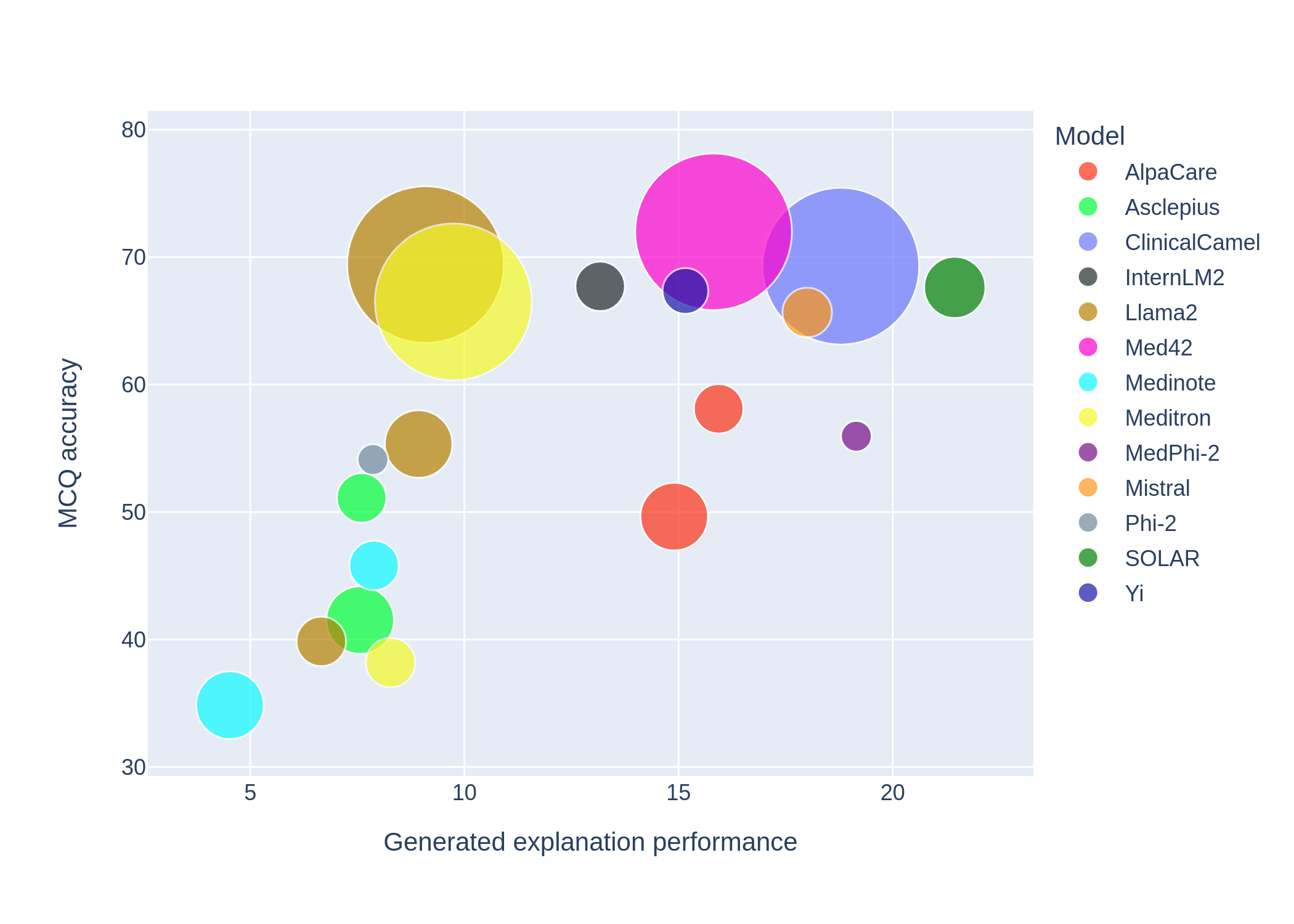}
    \caption{Scatter plot of model performance. The Y-axis is the macro average of accuracy based on logits (Table \ref{tab:result_choice}). The X-axis is the average score of generated explanations (Table \ref{tab:result_exp}). The dot size is proportional to the model size.}
    \label{fig:scatter1}
\end{figure*}

\begin{table*}[]
\begin{center}
\begin{tabular}{ccccccc}
\hline
\textbf{Model}    & \textbf{Size (B)} & \textbf{ROUGE-L} & \textbf{METEOR} & \textbf{BLEU} & \textbf{BERTScore} & \textbf{AVG} \\ \hline
Medinote          & 13                & 1.88                                 & 2.79                                & 0.46                              & 12.96                                & 4.52                             \\
Llama2            & 7                 & 4.92                                 & 4.03                                & 0.16                              & 17.52                                & 6.66                             \\
Asclepius  & 13                & 6.12                                 & 6.12                                & 0.32                              & 17.70                                & 7.56                             \\
Asclepius  & 7                 & 6.07                                 & 5.61                                & 0.22                              & 18.48                                & 7.60                             \\
Phi-2    & 2.7      & 5.77                                 & 7.51                                & 1.76                              & 16.41                                & 7.86                             \\
Medinote          & 7                 & 4.78                                 & 7.82                                & 2.14                              & 16.81                                & 7.89                             \\
Meditron          & 7                 & 5.15                                 & 7.96                                & 2.56                              & 17.43                                & 8.27                             \\
Llama2            & 13                & 6.65                                 & 6.89                                & 1.37                              & 20.80                                & 8.93                             \\
Llama2            & 70                & 6.41                                 & 6.71                                & 1.40                              & 21.84                                & 9.09                             \\
Meditron          & 70                & 7.42                                 & 8.32                                & 1.63                              & 21.59                                & 9.74                             \\
InternLM2         & 7                 & 10.30                                & 12.20                               & 3.89                              & 26.28                                & 13.17                            \\
AlpaCare   & 13                & 11.56                                & 11.97                               & 2.77                              & 33.29                                & 14.90                            \\
Yi                & 6                 & 10.97                                & 13.25                               & 4.79                              & 31.62                                & 15.16                            \\
Med42             & 70                & 11.03                                & 12.88                               & 3.46                              & 35.89                                & 15.82                            \\
AlpaCare   & 7                 & 12.43                                & 14.19                               & 3.64                              & 33.47                                & 15.94                            \\
Mistral           & 7                 & 12.59                                & 17.49                               & 5.28                              & 36.66                                & 18.00                            \\
ClinicalCamel     & 70                & 13.45                                & 17.38                               & 5.52                              & 38.80                                & 18.79                            \\
MedPhi-2 & 2.7      & 15.26                                & 17.75                               & 6.13                              & 37.45                                & 19.15                            \\
SOLAR             & 10.7              & 16.45                                & 20.17                               & 6.72                              & 42.46                                & 21.45                            \\
GPT3.5\_1106      & -                 & 21.71                                & 25.99                               & 14.07                             & 46.59                                & 27.09                            \\
GPT4\_1106        & -                 & 23.08                                & \textbf{35.74}                      & 14.40                             & \textbf{54.50}                       & 31.93                            \\
GPT4\_0125        & -                 & \textbf{24.83}                       & 35.21                               & \textbf{16.71}                    & 54.40                                & \textbf{32.79}                   \\ \hline
\end{tabular}
\caption{\label{tab:result_exp}
Explanation Generation performance (average across the 5 subjects for each evaluation metric).
}
\end{center}
\end{table*}

\subsection{Classification Accuracy - Logits}
Table \ref{tab:result_choice} shows the detailed results of all models. As expected, smaller language models demonstrated lower accuracy across specialties than larger models. Med42 showed the best overall performance. It showed outstanding performance in Biomedical Engineering and Clinical Laboratory Science (83.2\% and 84.9\% respectively). It performed on par with Meditron-70B in Clinical Psychology (84.9\%). In Occupational Therapy, Llama2-70B showed the highest accuracy (80.4\%). All models underperformed in Speech Language Pathology, with SOLAR performing the best (33.1\%). 

The effect of continued training is observed only in some models. MedPhi-2 demonstrated better performance than Phi-2, and this improvement was also found in AlpaCare-13B compared to Llama2-13B and Med42 compared to Llama2-70B. However, ClinicalCamel and Meditron-70B performed worse than Llama2-70B. This drop in performance could be due to task-specific challenges as some models may not effectively handle varied levels of specificity in MedExQA. 


\subsection{Classification Accuracy - Chat}
\label{sec:chatacc}
Classification accuracy using chat decreased in most of the models (Table \ref{tab:result_choice}). Phi-2, Llama2-13B, Yi, InternLM2, and Meditron-70B did not pass the pass mark indicating these models are not robust. Meditron-70B showed the biggest performance drop by 29.3\%. Llama2-70B also showed a significant performance drop in this testing by 28.5\%, although it passed in Biomedical Engineering. Of the 70B models we tested, ClinicalCamel was the most robust model (7.7\% decrease), and it scored higher than Med42 by 0.7\%.

Our model, MedPhi-2 was the most robust model among the passed ones (0.2\% decrease), and it outperformed AlpaCare-13B, Meditron-70B, Llama2-70B. This result highlights the importance of the supervised finetuning with in-domain instructions of high quality as more robust models, such as AlpaCare, ClinicalCamel, and MedPhi-2, were instruction-tuned with medical domain data, while Meditron-70B was just further pretrained.

GPT4\_0125, GPT4\_1106, and GPT3.5\_1106 outperformed all the open-source models. Even with the addition of high-performing closed-source models, there is still a universal failure in performance for Speech Language Pathology.

\subsection{Combining Classification Accuracy with Generated Explanation Performance}
Figure \ref{fig:scatter1} shows the relationship between model size and accuracy achieved in both MCQ (using logits) and generation performance. Generally, larger models tend to exhibit better performance as 70B models perform better than most of the other smaller models. However, SOLAR, Yi, and Mistral stand out as these smaller general domain models demonstrate competitive performance to the 70B medical LLMs. Further training on these foundation models holds great promise as we have seen with the Phi-2 model.

All medical LLMs with 13B (AlpaCare, Asclepius, and Meditron) exhibit worse performance in both MCQ accuracy and generation performance compared to their 7B counterparts. In fact, \textbf{Medinote-13B} is the worst-performing model. Also, 70B models do not always perform better than smaller models as Meditron-70B and Llama2-70B performed worse than many smaller models including AlpaCare and our model in the generation of reasonable explanations. 

The performance evaluation presented in Table \ref{tab:result_exp} also provides valuable insights into the efficacy of various models in generating explanations. Among the models evaluated, our model, \textbf{MedPhi-2} stands out in generating reasonable explanations as it outperformed all medical LLMs including 70B models. This result confirms the findings of Section \ref{sec:chatacc} which highlighted the importance of supervised finetuning with in-domain instructions.

The SOLAR model performed the best among the open-source models, suggesting its competitive capability in explanation generation although it was not trained specifically for the medical domain. However, even this best-performing open-source model demonstrates a significant performance gap (5.64) compared to the worst-performing closed-source model, GPT3.5\_1106, indicating the substantial advancements in OpenAI's GPT models.

\begin{figure*}[htbp]
    \centering
    \includegraphics[trim={0cm 0.1cm 0cm 0.1cm},clip,width=0.92\textwidth]{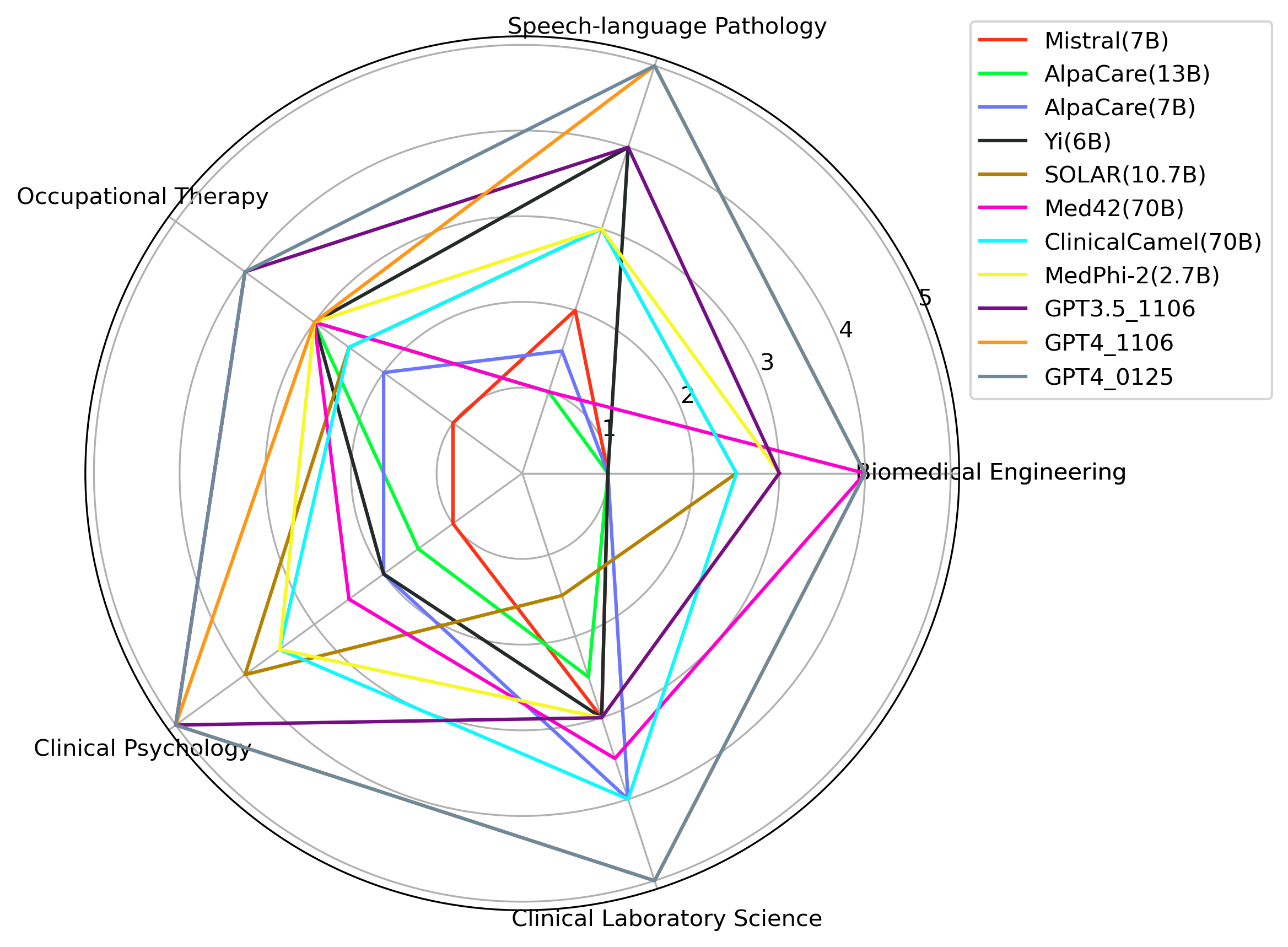}
    \caption{Human evaluation on the generated explanations, which scales from 0 to 5. The models in the legend are ordered by macro average from lowest to highest. Only models passed (3 or above) in at least one of the specialties are included.}
    \label{fig:radar_human}
\end{figure*}

Interestingly, despite the recent release of GPT4, the performance varies across different evaluation metrics. While the most recent release outperforms GPT4\_1106 on average, GPT4\_1106 still shows superior performance in METEOR and BERTScore. This highlights the importance of considering multiple metrics and nuances in model performance assessment, as different models may excel in distinct aspects of explanation generation.

\subsection{Evaluation - Human Evaluation}

Human evaluation of generated responses reveals that MedPhi-2 has the best quality among the open-source models (Figure \ref{fig:radar_human}). Our model was the only open-source model that passed (a score of 3 or above) in all specialties in MedExQA. In fact, \textbf{MedPhi-2} on par with \textbf{GPT3.5\_1106} in Biomedical Engineering and Clinical Laboratory Science, and with \textbf{GPT4\_1106} in Occupational Therapy.

The performance of models in Speech Language Pathology during human evaluation was relatively decent, which contrasts with results obtained through other evaluation methods. Appendix Figure \ref{fig:exp} provides an example of generated responses of the models, in the context of Speech-Language Pathology questions. \textbf{MedPhi-2} and \textbf{GPT3.5\_1106} generated the most coherent and accurate responses. However, other models generated irrelevant sentences or failed to provide explanations. \textbf{Medinote-13B} generated a case study example instead of answering the question and providing an explanation and \textbf{Asclepius-13B} hallucinated and provided an option for the answer that was not present and generated further incorrect explanations. Appendix Table \ref{tab:result_human} shows the detailed results. 

\subsection{Effect of additional explanation}

The effect of adding additional explanation was confirmed by analyzing the Pearson correlation between human evaluation and generation performance. When we used just one set of explanations the correlation was 0.9347, and this correlation increased to 0.9385 when we used two versions of explanations together. Although, the increase is small, this finding still indicates generation evaluation with multiple explanations aligns better with human evaluation, which is usually treated as the gold standard.

\section{Conclusion}
\label{sec:bibtex}

Our MedExQA benchmark proposes an effective methodology for evaluating LLMs beyond classification accuracy which can be used to assess the explainability of medical LLMs. While, the findings reveal that the generation of coherent and accurate explanations remains a challenging frontier for the current medical LLMs, the results also highlight an opportunity for a more robust automated comprehension assessment for LLMs because generation evaluation with multiple explanations aligned better with human assessment.

We also find that the ‘Speech Language Pathology’ dataset posed challenges for all language models, including GPT4. Speech Language Pathology could potentially be attributed to several factors, with one prominent explanation being the absence of relevant text in the corpora used to train the foundation model. As Speech Language Pathology is a highly specialized field that encompasses a wide range of topics related to rare diseases or disorders of speech and language, the collection of high-quality text for this specialty can be very challenging. However, it is important to acknowledge that confirming this hypothesis definitively poses a challenge due to the proprietary nature of the pretraining corpora used for training LLMs.

Through the development and evaluation of our MedPhi-2 model, we underscore the importance of targeted pretraining and fine-tuning strategies in improving explanation quality. The model showed the significant potential of LLMs in enhancing medical QA with explanations. Our benchmark and model will set the foundation for future advancements in medical research by facilitating the development and evaluation of medical LLMs.

\section*{Limitation}
\label{sec:limitation}
While MedExQA provides a robust benchmark for evaluating LLMs in the context of the medical domain, the current version only tests the model's ability in QA task, limiting its applicability in real-world clinical scenarios to a few applications. This limitation results from the manual collection process. Future work will extend our benchmark to include tasks such as summarizing clinical notes with accompanying explanations.

Though we performed the human evaluation of generated explanations of different LLMs through three authors, we performed this at a small scale, at 5 samples per specialty. Future work will seek to increase both the volume of samples and the number of annotators to provide a more robust method of assessing models’ performance. 


\section*{Broader Impacts and Ethics Statement}
\label{sec:ethics}
We release MedExQa under a Creative Commons Attribution-Non Commercial-ShareAlike 4.0 International License. MedPhi-2 follows the MIT license as it is based on Phi-2. License and copyright information and Terms of Use will be shared when the dataset and model are released. The dataset may be used for non-commercial purposes and any models trained using the dataset should be used only for research purposes.

Our work does not raise any major ethical concerns. All LLMs tested, including Phi-2, were used for research purposes only. While MedPhi-2 outperformed all medical variants of Llama2 models in generating accurate medical answers and explanations, MedPhi-2 is not rigorously tested for use in real-world clinical applications or scenarios. Thus, MedPhi-2 is not suitable for use in the clinical decision making process. This restriction of usage in clinical care is to mitigate any potential risks or harms such as wrong decisions from hallucinations which can lead to unwanted situations.

\clearpage
\bibliography{acl_latex_v2}

\clearpage
\appendix
\onecolumn
\section*{Appendix}
\label{sec:appendix}
\begin{figure*}[htbp]
\vskip -0.1in
    \centering
    \includegraphics[trim={2cm 3cm 3cm 3cm},clip,width=0.93\textwidth]{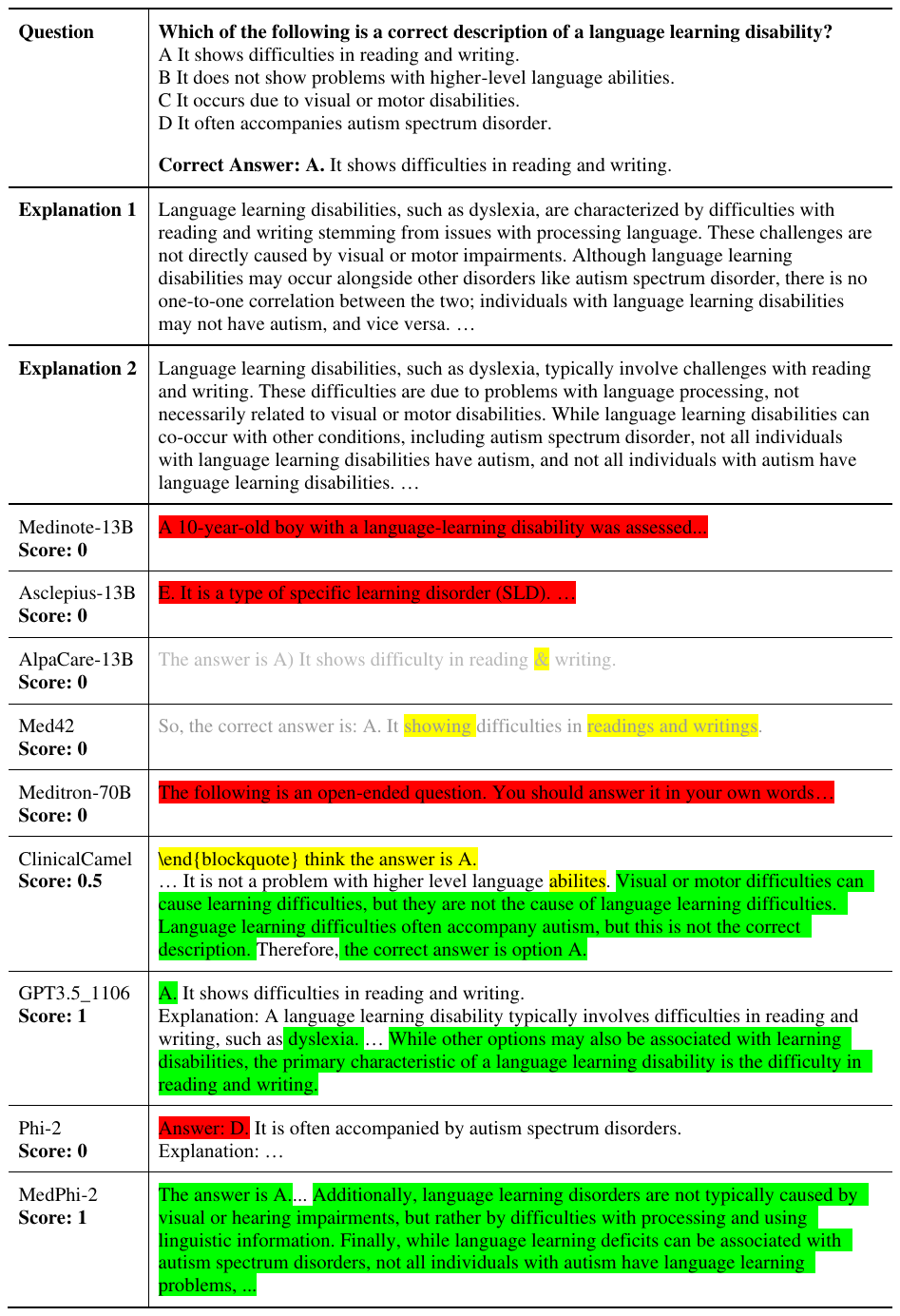}
    \caption{Example of data in Speech Language Pathology and Qualitative Analysis Example. Two sets of explanations, Explanation 1 and Explanation 2, are provided. The score given by humans is provided beneath the model name. The response with no explanations has a grey font color. Red shows the irrelevant or wrong sentences or phrases. Yellow demonstrates incoherent phrases or errors. Green highlights coherent and correct sentences.}
    \label{fig:exp}
\vskip -2.8in
\end{figure*}

\clearpage

\subsection{Frequency Plots}

\begin{figure}[htbp] 
    \centering
    \includegraphics[width=1.0\textwidth]{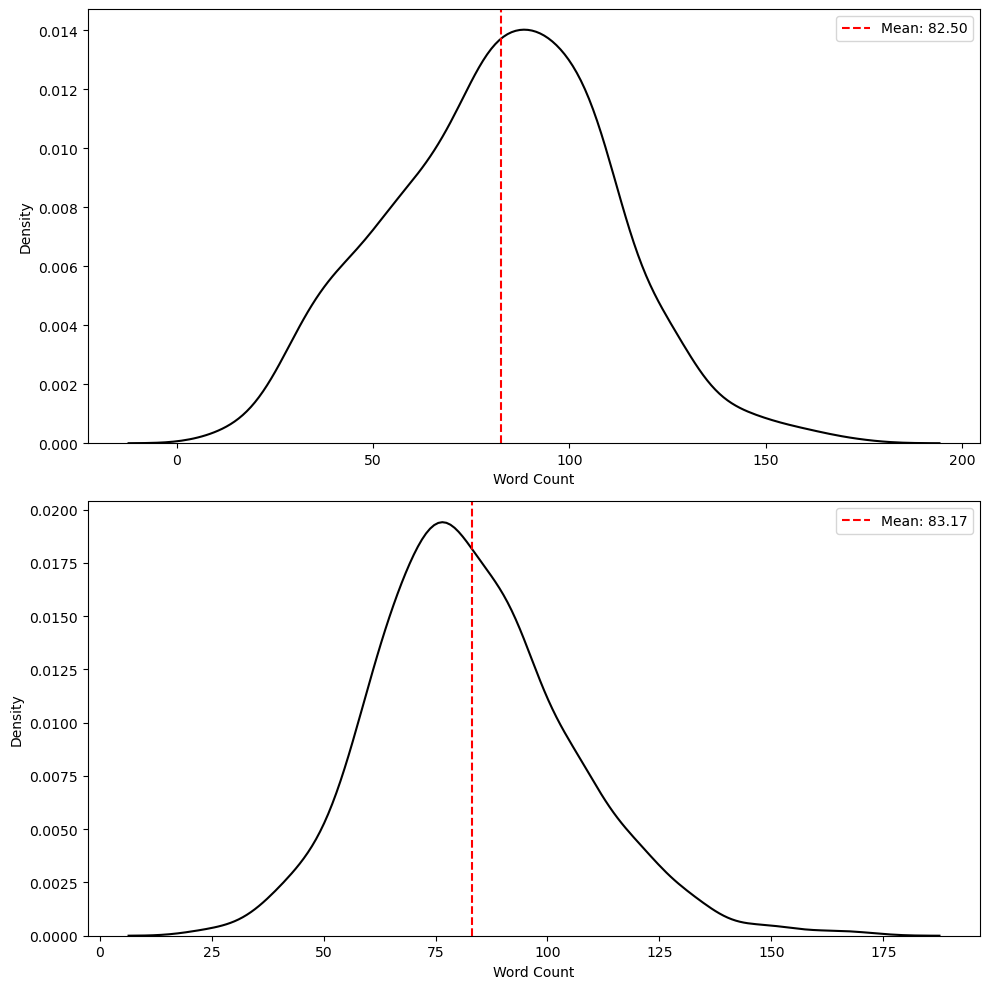}
    \caption{Word Count Distribution Plots for Explanations. Top: Explanation 1. Bottom: Explanation 2.}
    \label{fig:frequency-medexqa-explanations}
\end{figure}

We use frequency plots to demonstrate the word count distribution for the two different types of explanations in each dataset. For MedExQA, as shown in Figure~\ref{fig:frequency-medexqa-explanations}, the average length of the words in the first set of explanations is 82.50 and in the second set of explanations, is 83.17.

\subsection{First Three Words Frequency}


\begin{figure}[htbp]
    \begin{minipage}{0.5\linewidth}
        \centering
        \includegraphics[trim={9cm 8.5cm 7cm 6cm},clip,width=1.0\textwidth]{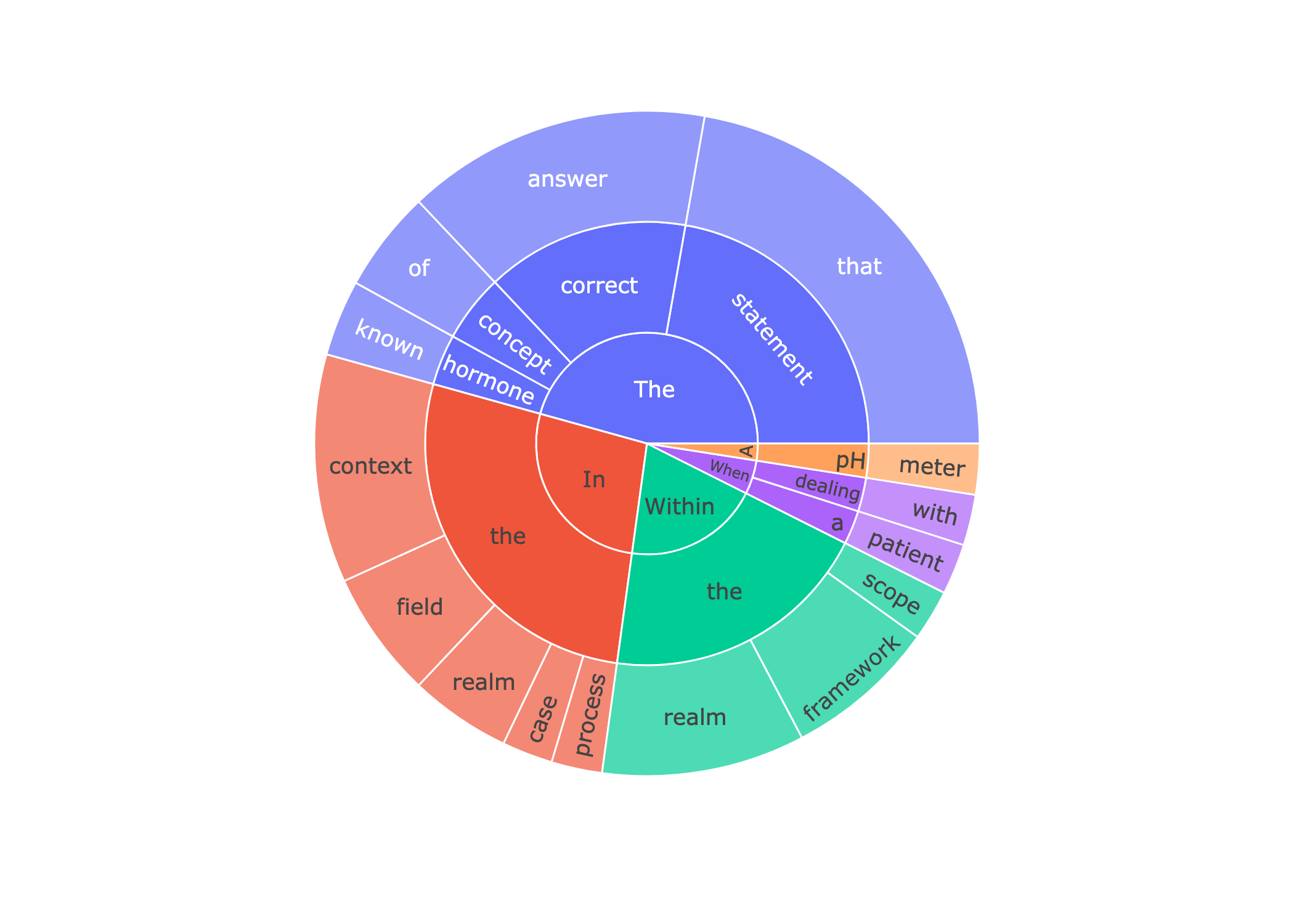}
    \end{minipage}
    \begin{minipage}{0.5\linewidth}
        \centering
        \includegraphics[trim={9cm 8.5cm 7cm 6cm},clip,width=1.0\textwidth]{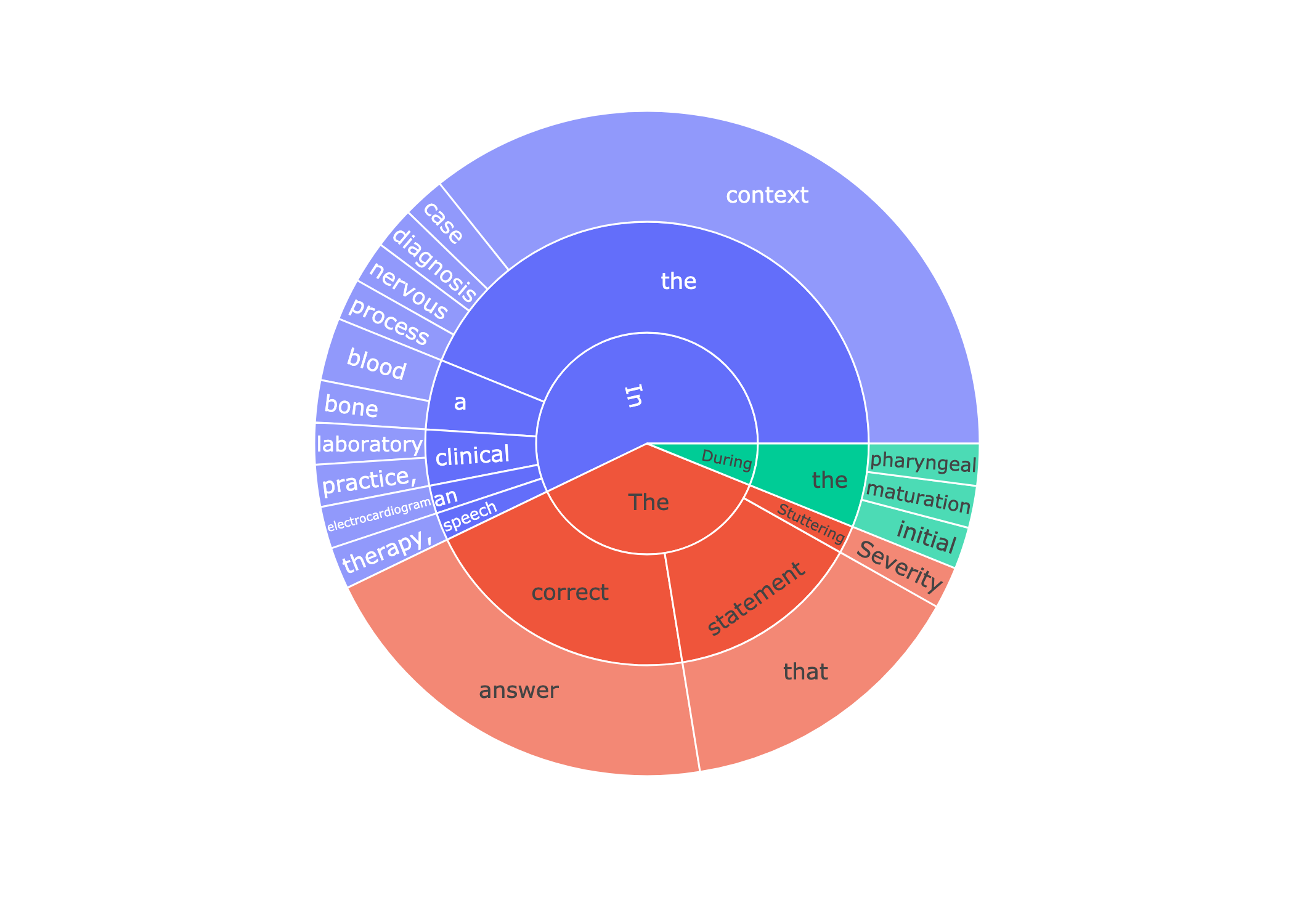}
    \end{minipage}
        \caption{First three words combination of explanations. Left: Explanation 1. Right: Explanation 2.}
        \label{fig:exp2}
\end{figure}

In Figure \ref{fig:exp2}, we present a detailed visualization of the lexical distribution within two distinct explanations from MedExQA datasets. For each pie chart, we combined the explanations from all five specialties. The pie chart encapsulates the hierarchical structure of the explanations, segmented into three concentric circles that correspond to the first, second, and third words of explanation, respectively. The top pie chart represents the word combination from explanation version 1, and the bottom pie chart represents the explanation version 2.

Upon examination, we note a convergence in linguistic choices, evidenced by recurring phrases such as "In the context" and "The correct answer." These phrases serve as linguistic anchors, providing a structured starting point for explanations. Despite this lexical overlap, the majority of the word choices exhibit significant variability. Some examples of this variability are "A pH meter" marked as orange and "When a patient" marked as purple on the top pie chart. By employing two versions of explanations that are semantically aligned yet lexically distinct, we aim to conduct a more holistic assessment of the model's generative outputs. 

\subsection{Baseline Models}

\subsubsection{Llama2 variants}
\textbf{Llama2} We use Llama2 Hugging Face weights released on the Hugging Face model repository\footnote{https://huggingface.co/meta-llama}. 7B, 13B, and 70B models without chat optimization are used in this work to assess the effect of continued training of the following Llama2 medical models with medical domain text. These models are trained on 2 trillion pretraining tokens in the general domain and have a context length of 4,096.

\textbf{ClinicalCamel} We use ClinicalCamel 70B weights from the Hugging Face model repository. It is a finetuned \textbf{Llama2-70B} model with instruction-tuning datasets made from medical articles and MedQA. It uses QLoRA for finetuning. The instruction tuning datasets are not released.

\textbf{Asclepius} We use Asclepius Llama2 weights released on the Hugging Face model repository. We use both 7B and 13B models which are further finetuned Llama2 models using instruction tuning dataset made from synthetic clinical notes. The synthetic clinical notes are generated from PMC-patients using GPT3.5 and turned into instruction-tuning datasets using GPT3.5. The synthetic clinical notes are used due to the privacy concerns of the real clinical notes. This training dataset is released.

\textbf{Med42} We use Med42 70B weights from the Hugging Face model repository. The details of the training dataset and training method are not available. The only detail available is that it was continued trained \textbf{Llama2-70B} model with medical domain text.

\textbf{AlpaCare} We use AlpaCare Llama2 weights from the Hugging Face model repository. Llama2 7B and 13B models were further finetuned on a medical self-instruct dataset made from the clinical seed set. The dataset is released.

\textbf{Meditron} We use Meditron weights released on the Hugging Face model repository. Both 7B and 70B models are used in this work. Meditron models are continued pretrained with clinical guidelines, medical articles abstracts, and full text of the articles. A subset of clinical guidelines are released.

\textbf{Medinote} We use Medinote weights released on the Hugging Face model repository. Both 7B and 13B models are used in this work. These models are further finetuned from the Meditron models to generate clinical notes from doctor and patient dialogues. Their training dataset is a synthetic dialog generated with ChatGPT from PMC-patients data.

\subsubsection{Other baseline models}

We extended our baseline models to other general domain baseline models with various sizes. 

\textbf{Mistral} We use Mistral-7B-v0.1 weight released on the Hugging Face model repository. The details of the training dataset remain unknown. However, this model is known to use Grouped Query Attention, which \textbf{Llama2-70B} also uses, and Sliding Window Attention. The model size is known to be 7.24B parameters, and this is slightly larger than \textbf{Llama2-7B}, 6.74B.

\textbf{Yi} We use Yi-6B weight released on the Hugging Face model repository. The model is trained on 3 trillion pretraining tokens in the general domain and has a context length of 4,096. The model size is known to be 6.06B parameters, which is smaller than other 7B models.

\textbf{Phi-2} We use Phi-2 model weight released on the Hugging Face model repository. It has 2.78B parameters and is trained on the augmented textbook corpus, 1.4 trillion tokens. This is the smallest model in our paper.

\textbf{SOLAR} We use SOLAR-10.7B-v1.0 model weight released on the Hugging Face model repository. The model size is 10.7 billion parameters. It uses depth-wise scaling called Depth up-scaling and continued pretraining of the scaled model. However, the pretraining dataset details are unknown.

\textbf{InternLM2} We use InternLM2-7b model weight from the Hugging Face model repository. The details of the training method and data are unknown.

\subsection{Result Tables}

\begin{table*}[bh]
\begin{center}
\begin{tabular}{cccccccc}
\hline
\textbf{Model}   & \textbf{Size (B)} & \textbf{BE} & \textbf{CP} & \textbf{SLP} & \textbf{OT} & \textbf{CLS} & \textbf{AVG} \\ \hline
Llama2           & 13                & 0                              & 0                                & 0                                      & 0                                 & 0                                        & 0            \\
Meditron         & 70                & 0                              & 0                                & 0.5                                    & 0.5                               & 0.5                                      & 0.3          \\
Asclepius & 13                & 0                              & 1.5                              & 0                                      & 0                                 & 1                                        & 0.5          \\
Medinote         & 13                & 0.5                            & 0.5                              & 0.5                                    & 0                                 & 1                                        & 0.5          \\
Meditron         & 7                 & 0.5                            & 0                                & 1                                      & 1                                 & 0                                        & 0.5          \\
Llama2           & 7                 & 0                              & 1                                & 0                                      & 1.5                               & 1                                        & 0.7          \\
Llama2           & 70                & 0.5                            & 0                                & 1                                      & 0.5                               & 2                                        & 0.8          \\
Asclepius & 7                 & 1                              & 1.5                              & 0                                      & 0                                 & 2                                        & 0.9          \\
Medinote         & 7                 & 1.5                            & 0.5                              & 1                                      & 0                                 & 1.5                                      & 0.9          \\
InternLM2        & 7                 & 2                              & 2                                & 1                                      & 0                                 & 1.5                                      & 1.3          \\
Phi-2            & 2.7               & 2                              & 2                                & 0                                      & 1                                 & 2                                        & 1.4          \\
Mistral          & 7                 & 1                              & 1                                & 2                                      & 1                                 & 3                                        & 1.6          \\
AlpaCare  & 13                & 1                              & 1.5                              & 1                                      & 3                                 & 2.5                                      & 1.8          \\
AlpaCare  & 7                 & 1                              & 2                                & 1.5                                    & 2                                 & 4                                        & 2.1          \\
Yi               & 6                 & 1                              & 2                                & 4                                      & 3                                 & 3                                        & 2.6          \\
SOLAR            & 10.7              & 2.5                            & 4                                & 3                                      & 2.5                               & 1.5                                      & 2.7          \\
Med42            & 70                & 4                              & 2.5                              & 1                                      & 3                                 & 3.5                                      & 2.8          \\
ClinicalCamel    & 70                & 2.5                            & 3.5                              & 3                                      & 2.5                               & 4                                        & 3.1          \\
MedPhi-2         & 2.7               & 3                              & 3.5                              & 3                                      & 3                                 & 3                                        & 3.1          \\
GPT3.5\_1106     & -                 & 3                              & 5                                & 4                                      & 4                                 & 3                                        & 3.8          \\
GPT4\_1106       & -                 & 4                              & 5                                & 5                                      & 3                                 & 5                                        & 4.4          \\
GPT4\_0125       & -                 & \textbf{4}                     & \textbf{5}                       & \textbf{5}                             & \textbf{4}                        & \textbf{5}                               & \textbf{4.6} \\ \hline
\end{tabular}
\caption{\label{tab:result_human}
Explanation Generation performance (human evaluation). "BE": Biomedical Engineering; "CP": Clinical Psychology; "SLP":	Speech Language Pathology; "OT":	Occupational Therapy;	"CLS": Clinical Laboratory Science; "AVG": Average score.
}
\end{center}
\end{table*}




\end{document}